\documentclass{IEEEtran}

\usepackage[utf8]{inputenc} 
\usepackage[T1]{fontenc}    
\usepackage{hyperref}       
\usepackage{url}            
\usepackage{booktabs}       
\usepackage{amsfonts}       
\usepackage{amsmath}
\usepackage{nicefrac}       
\usepackage{microtype}      
\usepackage{float}
\usepackage[pdftex]{graphicx} 
\usepackage{color}
\usepackage{amssymb}
\usepackage{bm}
\usepackage{multirow}
\usepackage{svg}
\usepackage{graphicx}
\usepackage{subfig}
\usepackage{algorithm2e}
\usepackage[font=small,labelfont=bf]{caption}
\usepackage{algpseudocode}

\usepackage[shortcuts]{extdash}

\usepackage[textsize=tiny,textwidth=1.2cm]{todonotes}

\hypersetup{hidelinks}

\def\tablename{Table}

\begin{document}

\title{Streaming convolutional neural networks for end-to-end learning with multi-megapixel images}

\author{
  Hans Pinckaers\textsuperscript{*},
  Bram van Ginneken,
  Geert Litjens
\begin{footnotesize}
  \begin{center}
  Computational Pathology Group, \\ Radboud University Medical Center, \\ The Netherlands \\
  \textsuperscript{*}hans.pinckaers@radboudumc.nl
  \end{center}
\end{footnotesize}
}

\maketitle

\begin{abstract}
Due to memory constraints on current hardware, most convolution neural networks (CNN) are trained on sub\=/megapixel images. For example, most popular datasets in computer vision contain images much less than a megapixel in size (0.09MP for ImageNet and 0.001MP for CIFAR\=/10). In some domains such as medical imaging, multi-megapixel images are needed to identify the presence of disease accurately. We propose a novel method to directly train convolutional neural networks using any input image size end\=/to\=/end. This method exploits the locality of most operations in modern convolutional neural networks by performing the forward and backward pass on smaller tiles of the image. In this work, we show a proof of concept using images of up to 66\=/megapixels (8192$\times$8192), saving approximately 50GB of memory per image. Using two public challenge datasets, we demonstrate that CNNs can learn to extract relevant information from these large images and benefit from increasing resolution. We improved the area under the receiver\=/operating characteristic curve from 0.580 (4MP) to 0.706 (66MP) for metastasis detection in breast cancer (CAMELYON17). We also obtained a Spearman correlation metric approaching state\=/of\=/the\=/art performance on the TUPAC16 dataset, from 0.485 (1MP) to 0.570 (16MP). Code to reproduce a subset of the experiments is available at \url{https://github.com/DIAGNijmegen/StreamingCNN}.
\end{abstract}

\section{Introduction}

Convolutional neural networks (CNN) are the current state\=/of\=/the\=/art machine learning algorithms for many tasks in computer vision, such as classification or segmentation. Ever since Krizhevsky et al. won ImageNet\cite{Russakovsky2015} with a CNN\cite{Krizhevsky2012} in 2012, these networks have become deeper\cite{He2016} and wider\cite{Zagoruyko2016} to further improve accuracy. Training these larger networks requires large amounts of computer memory, which increases exponentially with increasing image size. To avoid shortcomings in memory, most natural image datasets in computer vision contain sub\=/megapixel images: 0.09 megapixel for ImageNet\cite{Russakovsky2015} and 0.001 megapixel for CIFAR\=/10\cite{Krizhevsky2009}. In several domains such as remote sensing or medical imaging, there is a need for training CNNs with multi\=/megapixel\=/sized images -- containing both global contextual and local textural information -- to obtain accurate models.

Computer memory becomes a limiting factor because the conventional backpropagation algorithm for optimizing deep neural networks requires the storage of intermediate activations. Since the size of these intermediate activations in a convolutional neural network increases proportionate to the input size, memory quickly fills up with images of multiple megapixels. As such, only small CNNs could be trained with these images and state\=/of\=/the\=/art architectures would be out of reach, even on large computing clusters.

In this paper, we propose a novel method to directly train state\=/of\=/the\=/art convolutional neural networks using any input image size end\=/to\=/end. This method exploits the locality of most operations in modern convolutional neural networks by tiling the forward and backward pass in combination with gradient checkpointing. We first empirically established equivalence between our tile\=/based approach and an unmodified convolutional neural network on a subset of ImageNet, ImageNette\cite{Howard2019}. Then we applied this method to two public datasets: the CAMELYON17 dataset\cite{Litjens2018} for metastases detection in lymph nodes, and the TUPAC16 dataset\cite{Veta2019} for predicting a proliferation score based on gene expression. In both cases, task\=/specific performance increased with larger input image sizes.

\section{Related work}
Several authors have suggested approaches to train convolutional neural networks (CNNs) with large input images while preventing memory bottlenecks. Their methods can be roughly grouped into three categories: (A) altering the dataset, (B) altering usage of the dataset, and (C) altering the network or underlying implementations.

\subsection{Altering the dataset}
If images are too large to fit in the memory of the processing unit, we could downsample the image or divide the image into smaller parts, i.e., patches. The latter approach has been prevalent in both remote sensing and medical imaging\cite{Ma2019, Litjens2017}. However, both approaches have significant drawbacks: the former results in a loss of local details, whereas the latter results in losing global contextual information.

The common approach of training on patches typically involves creating labels for every patch, which can be time- and cost\=/intensive. It is sometimes not even possible to produce patch\=/level labels: if a hypothetical task is to predict whether an aerial image shows a city or a village, it is impossible to create informative labels for individual patches only containing several houses. In the Kaggle Data Science Bowl 2017\cite{Kaggle2017}, participants were asked to classify chest CT images as containing lung cancer or not. In this case, a non\=/expert could not label individual 3D patches of the CT scan without knowing the location of cancer.

\subsection{Altering usage of the dataset}
When we can assume that individual patches contain enough information to predict the image\=/level label, the classification can be formalized under the classic multiple\=/instance\=/learning (MIL) paradigm. In MIL, each image is considered a bag consisting of patches where a positive bag has at least one positive patch and a negative bag none. In the deep learning case, a model is trained in a weakly supervised manner on patches, where the patch with the highest predicted probability is used for backpropagation. Other approaches involve taking the average of the patch predictions or a learned weighted average from low\=/dimensional patch embeddings\cite{Quellec2017,Ilse2018}.

In this approach, the receptive field of a network is always at most the size of the patch. The model disregards spatial relationships between patches, limiting the incorporation of contextual information.

By first learning to decide which regions should be analyzed at a higher resolution,  the problem that a full image cannot be used can also be circumvented\cite{Dong2018, Mnih2014, Katharopoulos2019, Recasens2018}. These methods rely on the existence of clues in downsampled images to guide the analysis to informative higher resolution patches, which might not be present. Additionally, for analysis of the selected region, these methods si till use patch\=/based analysis with the same caveats as mentioned before.

Another way to utilize datasets with large images is proposed by Tellez et al.\cite{Tellez2019}. To compress the image to a lower\=/dimensional space, they proposed unsupervised learning. The model is trained patch\=/by\=/patch to reconstruct the original patch. An intermediate feature map of the model (i.e., the embedding) can subsequently be used as a lower\=/dimensional representation per patch. After training, the whole image is compressed patch\=/by\=/patch. A model is subsequently trained on these embeddings, having the receptive field of the whole image while requiring less memory.

Since the compression network is trained by reconstruction, the same compression network can be used for different tasks. However, this means that the low\=/dimensional embedding is not meant for a specific task and may have compressed away useful information. Our approach involves one network which learns to compress task\=/relevant information.

\subsection{Altering the network or underlying implementations}
The memory bottleneck can also be circumvented with memory\=/efficient architectures or memory\=/efficient implementations of existing architectures. Recently, Gomez et al.\cite{Gomez2017} published a method to train deep residual neural networks using less memory, termed the Reversible Residual Network. With these networks, some layer activations are recomputed from others on demand, reducing the total memory required. Network architectures can also be altered to utilize cheaper computational operation, such as depthwise separable convolutions\cite{Chollet2017} or fewer parameters\cite{Tan2019}. Our method does not require reducing the number of parameters and works with most types of layers. Another method to reduce memory usage is to recover intermediate activations by doing partial forward passes during backpropagation, termed gradient checkpointing\cite{Chen2016a}. This method is similar to our approach, but the whole activation feature map of some layers still need to be stored in memory, limiting the use of multi\=/megapixel images. 

Another memory\=/saving approach is to share memory between tensors with duplicate or recomputable values\cite{Huang2017, Bulo2018}, to develop neural networks with reduced precision using half\=/precision or mixed precision\cite{Vanhoucke2011}, or to swap data between random access memory (RAM) and graphics processing unit (GPU) memory\cite{Zhang2019}. These methods are usually insufficient for training with large multi\=/megapixel images; our proposed method can work orthogonally to them.

\section{Methods}

To achieve our goal of training CNNs with multi\=/megapixel images, we significantly reduce the memory requirements. Memory demand is typically highest in the first few layers of state\=/of\=/the\=/art CNNs before several pooling layers are applied because the intermediate activation maps are large. These activation maps require much less memory in subsequent layers. We propose to construct these later activations by streaming the input image through the CNN in a tiled fashion, changing the memory requirement of the CNN to be based on the size of the tile and not the input image. This method allows the processing of input images of any size.

Several problems arise when trying to reconstruct the later activation map tile\=/by\=/tile. Firstly, convolutional layers handle image borders in different ways, either by padding zeros to perform a ``same'' convolution or by reducing the image size to perform a ``valid'' convolution. Secondly, in tile\=/based processing, border effects occur at both the image borders and the tile borders; naive tiling of the input image would thus result in incomplete activation maps and gradients for backpropagation. Lastly, intermediate feature maps of the tiles still need to be stored in memory for backpropagation, which would counteract the streaming of tiles. We solve these problems by developing a principled method to calculate the required tile overlap throughout the network in both the forward and backward pass and by using gradient checkpointing.

\begin{figure*} 
    \centering
    {\includegraphics[width=\textwidth]{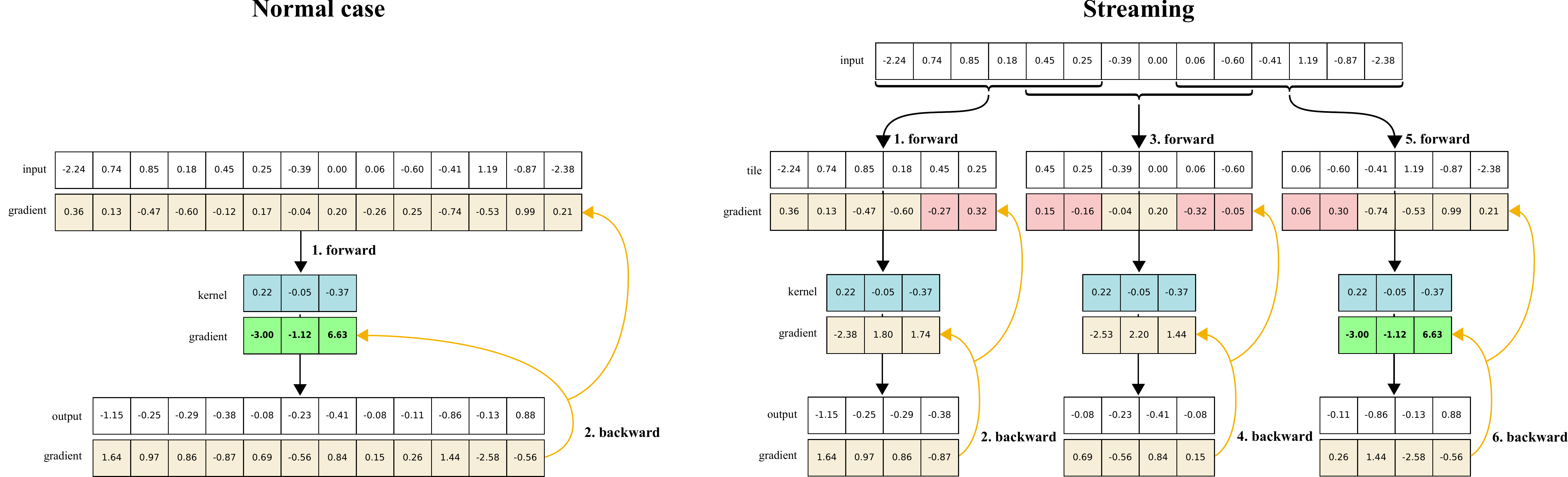}}%
    \caption{Schematic overview of streaming in a one\=/dimensional case, using a 1x3 kernel. We can calculate the same gradients with smaller input and output shapes, saving memory. Colored in red are incomplete gradients of the input tile, illustrating that if we chain two of these one\=/dimensional convolutions together, more overlap is needed (four values instead of the two pictured). During streaming, the gradients of the convolutional kernel are not reset between tiles and are summed with the previous gradients. After the three tiles have been backpropagated, the kernel gradients are equal to the normal case. Gradient checkpointing is also used to save additional memory, but is omitted here for clarity.}
    \label{figure:streamingSGD}
\end{figure*}

We first explain the reconstruction of the intermediate activation map in the forward pass in section \ref{forwardpass}, then describe the backward pass in section \ref{backwardpass}, elaborate on how to calculate the tile overlap in section \ref{calculateoverlap}, and finish with the limitations of this method in section \ref{limitations}. See Figure \ref{figure:streamingSGD} for a graphical representation of the method.

\subsection{Streaming during the forward pass}
\label{forwardpass}
Without loss of generality, we explain the method in the discrete one\=/dimensional case. Let us define $x \in \mathbb{R}^{N}$ as the one\=/dimensional real\=/valued vector with $N$ elements. In discrete one\=/dimensional space, a ``valid'' convolution\footnote{By convention we used the term \textit{convolution} although the mathematical operation implemented in most machine learning frameworks (e.g., TensorFlow, PyTorch) is a cross\=/correlation.} (*) with a kernel with $n$ weights $w \in \mathbb{R}^n$, and stride 1, is defined as:

\begin{equation} \label{eq:1}
(x * w)_k = \sum_{i=0}^{n} w_{i}x_{k+i}
\end{equation}

where $k \in \{0,\ldots,f\}$ and $f=N - n$, for any kernel with length $n \leq N$ (for clarity, we will start all indices from 0). Our goal is to decrease the memory load of an individual convolution by tiling the input.  Following \eqref{eq:1}, we can achieve the same result as $x * w$, by doing two convolutions on the input:
\begin{gather} 
a = \{(x * w)_0,\ldots,(x * w)_{f//2}\} \label{eq:2}\\
b = \{(x * w)_{f//2+1},\ldots,(x * w)_f\} \label{eq:3}
\end{gather}
where $//$ denotes a divide and floor operation.

By definition of concatenation ($\frown$):
\begin{equation} \label{eq:4}
\{(x * w)_0,\ldots,(x * w)_f\}= a \frown b
\end{equation}

To ensure that the concatenation of both tiles results in the same output as for the full vector, we need to increase the size of the tiles, resulting $o=n-1$ overlapping values. The values $\{ x_0,\ldots,x_{f//2+o} \}$ are required to calculate $a$, and $\{ x_{f//2+1-o},\ldots,x_N \}$ for $b$. 

Since the tiles are smaller than the original vector, these separate convolutions require less memory when executed in series. By increasing the number of tiles, memory requirements for individual convolution operations can be reduced even further.

Without loss of generality, the above can also be extended to multiple layers in succession including layers with stride $> 1$ (e.g., strided convolutions and pooling layers) which are also commonly used in state\=/of\=/the\=/art networks.

When one applies this tiling strategy naively, no memory benefit is obtained as each tile's intermediate activation would still be stored in memory to allow for backpropagation. We use gradient checkpointing to resolve this: We only store the activations after the concatenation of the tiles -- where the memory burden is small. This does require recalculation of all intermediate activations for all tiles during backpropagation, but again, only has a memory requirement of processing of a single tile. The trade\=/off between memory use and re\=/computation can be controlled through the selection of the concatenation point in the network.

From this point onward, the term \textit{streaming} refers to the tiling of a vector, applying kernel operations, and concatenating the results.




\RestyleAlgo{boxruled}
\LinesNumbered
\IncMargin{0.5em}
\begin{algorithm}[h]
\DontPrintSemicolon
\SetAlCapFnt{\small}
\SetAlCapNameFnt{\small}
\SetNlSty{textrm}{}{}
\SetAlgoNlRelativeSize{-3}
\SetAlgoCaptionLayout{captionwithmargin}
\caption{\em Forward and backward pass with streaming through the bottom layers of the network with tiles.}
\SetKwInOut{Input}{in}
\SetKwInOut{Output}{out}
\SetKwFor{With}{with}{}{end}
\small
\Indmm
    \Input{$n$ convolutional $layers$, $x$ image to stream with $m$ tiles $t$ at $coordinates$, and $i$ the last layer to stream. \texttt{crop\_unique} uses indices from \textit{Algorithm \ref{algorithm:crop}.}}
    \Output{$grads$ containing gradients per layer.}
\Indpp
    \BlankLine
$o\gets[]$ \Comment{array to collect tile outputs}\;
\With {no gradient computation}{
\For {$c$ in coordinates}{
    $t\gets$ \texttt{crop}$(x, c)$\;
    $o[c]\gets$ \texttt{forward}$(layers[0..i], t)$\;
}
}
$stream\_o\gets$ \texttt{concat}$(o[0..m])$\;
$pred\gets$ \texttt{forward}$(layers[i..n], stream\_o)$\;
$loss\gets$ \texttt{criterion}$(pred)$\;
$g\gets$ \texttt{backward}$(layers[n..i], loss)$\;
$filled\gets[]$ \Comment{array to remember backpropped indices}\;
\For {$c$ in coordinates}{
    $o[c]\gets$ \texttt{forward}$(layers[0..i], t)$ \Comment{checkpointing}\;
    $g_t \gets$ \texttt{crop\_relevant\_gradient}$(g, c)$\;
    \For {$l$ in $layers[i..0]$}{
        $g_t \gets$ \texttt{backward}$(l, g_t)$\;
        $o_u, g_u, filled[l]\gets$ \texttt{crop\_unique}$(l, o[c], c, filled[l])$\;
        $g_{kernel}\gets$ \texttt{backward}$(o_u, g_u)$\;
        $grads[l]\gets$ \texttt{sum\_gradients}$(g_{kernel}, grads[l])$\;
    }
}
\end{algorithm}
\DecMargin{0.5em}

\subsection{Streaming during backpropagation}
\label{backwardpass}
The backward pass of multiple convolutions can also be calculated by utilizing the tiles. To start, let us define $p$ as the output after streaming. The derivative of a weight in a convolutional kernel is defined as:

\begin{equation} 
\Delta w_j = \sum_{i=0}^{\lvert p \rvert-1}
\begin{cases}
  \Delta p_i x_{i+j}, & \text{if}\ i - j \geq 0 \text{ and } i - j < \lvert p \rvert \\
%
  0, & \text{otherwise}
\end{cases}
\end{equation} 
where $\lvert \cdot\rvert$ denotes the length of a vector.

While streaming, this sum has to be reconstructed through the summation of the gradients of all tiles, which will result in the same gradient again:

\begin{equation} \label{eq:6}
\Delta w_j = \sum_{i=0}^{\lvert a \rvert-1} \Delta a_i x_{i+j} + \sum_{i=0}^{\lvert b \rvert-1} \Delta b_i x_{i+j+f//2}
\end{equation}

The gradient of the input can be calculated with a similar sum, but then shifted by the kernel size:
\begin{equation} 
\Delta x_i = \sum_{j=0}^{n-1} 
\begin{cases}
  w_j\Delta p_{i-j}, & \text{if}\ i-j \geq 0 \text{ and } i - j < \lvert p \rvert  \\
  0, & \text{otherwise}
\end{cases}
\end{equation}

This formula is equal to a convolution with a flipped kernel $w$ on $\Delta p$ padded with $n - 1$ zeros (e.g., $flip(w) * [0, 0, \Delta p_1 ...  \Delta p_n, 0, 0]$, when $n=3$), often called a ``full'' convolution. Thus, analog to the forward pass, the backpropagation can also be streamed.

However, overlapping values of the output $p$ are required when streaming the backpropagation, similar to the overlapping values of the input $x$ required in the forward pass. To generate overlapping values for the output $p$, the overlap $o$ for the input $x$ needs to be increased to calculate the full $\Delta x$.\footnote{Zero\=/padding the tiles before the convolution does not help because these zeros do not exist in the original vector, hereby invalidating the gradients at the border as well.}


\subsection{Efficiently calculating required tile overlap for complex architectures}
\label{calculateoverlap}

Some recent state\=/of\=/the\=/art networks (e.g., ResNet and DenseNet) contain different paths through the network that sum or concatenate activations from different layers together. These paths make it difficult to manually calculate the required tile overlap for streaming.

To calculate the overlap for such networks, we temporarily replace all convolutional kernel parameters with $\frac{1}{n}$, where $n$ was the length of the kernel. This causes each entry in the convolutional layer's output to be the average of the input image spanned by the convolutional kernel. We then pass an all\=/ones tile through the network. The required overlap will be the number of non\=/maximum values in the activation maps and gradients at the border of the tiles, see Algorithm \ref{algorithm:crop}.

\RestyleAlgo{boxruled}
\LinesNumbered
\IncMargin{0.5em}
\begin{algorithm}[h]
\DontPrintSemicolon
\SetAlCapFnt{\small}
\SetAlCapNameFnt{\small}
\SetNlSty{textrm}{}{}
\SetAlgoNlRelativeSize{-3}
\SetAlgoCaptionLayout{captionwithmargin}
\caption{\em Finding which areas of streamed feature maps and gradients contain values equal to feature maps and gradients in an backpropagated full\=/resolution image.}
\label{algorithm:crop}
\small
\SetKwInOut{Input}{in}
\SetKwInOut{Output}{out}
\Indmm
    \Input{$t$ tile at the desired tile size, containing a constant value, $layers$ containing $n$ number of convolutional neural network layers, and $i$ the last layer to stream.}
    \Output{statistics for forward and backward pass $invalid\_forw\_i$, $invalid\_back\_i$, and $output\_stride$.}
\Indpp
    \BlankLine
    $output\_stride\gets 1$\;
    \For {$l$ in $layers[0..i]$}{
        $kernel\_backup_l\gets kernel_l$\;
        $kernel_l\gets 1/\texttt{len}(kernel_l)$\;
        $t\gets$ \texttt{forward$(l$, $t$)}\;
        $invalid\_forw\_i\gets$ \texttt{non\_max\_indices}$(o)$\;
        $output\_stride\gets output\_stride * l.stride$
    }
    $o\gets$ \texttt{forward}$(layers[i..n], o)$\;
    $loss\gets$ \texttt{criterion}$(o)$\;
    $g\gets$ \texttt{backward}$(layers[n..i], loss)$\;
    \For {$l$ in $layers[i..0]$}{
        $g\gets$ \texttt{backward}$(l, g)$\;
        $invalid\_back\_i\gets$ \texttt{non\_max\_indices}$(g)$\;
        $kernel_l\gets kernel\_backup_l$\;
    }
\end{algorithm}
\DecMargin{0.5em}


\subsection{Limitations}
\label{limitations}
With small tiles, the overlap can be a significant part of the tile, counteracting the memory gains. Since we leverage the method for high\=/resolution images using large tiles, the memory gains outweigh this overhead.

Furthermore, due to the use of gradient checkpointing, the method will perform multiple forward and backward operations to calculate intermediate activations. This results in longer processing time than it would take if the image could fit on the GPU (see Table \ref{tab:times}). The processing time increases almost linearly with the number of tiles.

Finally, since the method relies on the local properties of convolutions and pooling operations, trying to use other operations that break this locality will result in invalid results (e.g., operations that rely on all the feature map values such as BatchNormalization\cite{Ioffe2015}). Note that these operations can still be used in the non\=/streaming later layers of the network.

\begin{table}[h]
\footnotesize
\centering
\caption{Forward and backward pass time and peak memory usage for different input image sizes.}
\label{tab:times}
\begin{tabular}{lllll}
\toprule
\textbf{Input} & \textbf{Tile size (n)} &\textbf{Forward}& \textbf{Backward} & \textbf{Memory} \\
\midrule
1024\textsuperscript{2} & 1024\textsuperscript{2} (1) & 0.5 ms & 45 ms & 1957 MB \\
1024\textsuperscript{2} & 527\textsuperscript{2} (4) & 16 ms & 89 ms & 513 MB \\
1024\textsuperscript{2} & 358\textsuperscript{2} (9) & 22 ms& 111 ms & 331 MB \\
1024\textsuperscript{2} & 273\textsuperscript{2} (16) & 27 ms& 147 ms & 202 MB \\
\midrule
1024\textsuperscript{2} & 1024\textsuperscript{2} (1) & 0.5 ms & 45 ms & 1957 MB \\
2048\textsuperscript{2} & 1039\textsuperscript{2} (4) & 47 ms & 308 ms & 1994 MB \\
4096\textsuperscript{2} & 1039\textsuperscript{2} (16) & 187 ms & 1373 ms & 2283 MB \\
8192\textsuperscript{2} & 1039\textsuperscript{2} (64) & 753 ms & 5127 ms & 3435 MB \\
\bottomrule
\end{tabular}
\vspace{0.2cm}
\caption*{\textit{Notes:} Performance of three 2D\=/convolutional layers with respectively 3, 64, and 3 output channels and kernel\=/size 3 on an RTX 2080Ti GPU. The memory increase that can be appreciated with increasing input sizes is due to the size of the output gradient.}
\end{table}

\section{Evaluation}
We evaluated the streaming method with three different datasets and network architectures. First, in Section \ref{section:imagenette}, we evaluated whether a CNN using streaming trains equivalently to the conventional training. We trained the same CNN on a small subset of the ImageNet dataset, ImageNette, using both methods\cite{Howard2019}. Second, in Section \ref{section:tupac}, we evaluated the usage of streaming on a regression task in the public TUPAC16\cite{Veta2019} dataset with high\=/resolution images (multiple gigapixels) and only image\=/level labels which are based on the mean RNA expression of 11 proliferation\=/associated genes. We trained multiple networks using increasing image resolutions and network depth. Finally, in Section \ref{section:camyleon}, we evaluated streaming in a classification task using the image\=/level labels of the CAMELYON17 dataset\cite{Bandi2019}.

An open\=/source implementation of the streaming algorithm and the ImageNette experiments can be found at \url{https://github.com/DIAGNijmegen/StreamingCNN}.

\section{Experiments on ImageNette}
\label{section:imagenette}
To evaluate whether a neural network using streaming trains equivalently to the conventional training method, we trained a CNN on small images using both methods starting from the same initialization. We used a subset of the ImageNet dataset, ImageNette, using 100 examples of 10 ImageNet classes (tench, English springer, cassette player, chain saw, church, French horn, garbage truck, gas pump, golf ball, parachute), analog to \cite{Howard2019}.

\begin{table}[h]
\footnotesize
\centering
\caption{Network architecture for Imagenette experiment}
\label{tab:imagenetnet}
\begin{tabular}{lll}
\toprule
\textbf{Layers} & \textbf{Kernel size} & \textbf{Channels} \\
\midrule
2D convolution & 7x7 & 16 \\
2D max-pool & 2x2 & 16 \\
2D convolution & 3x3 & 32 \\
2D max-pool & 2x2 & 32 \\
2D convolution & 3x3 & 64 \\
2D max-pool & 2x2 & 64 \\
2D convolution & 3x3 & 128 \\
2D max-pool & 2x2 & 128 \\
2D convolution & 3x3 & 256 \\
2D max-pool & 10x10 & 256 \\
Fully connected & 10 \\
\bottomrule
\end{tabular}
\end{table}

\subsection{Data preparation}
We selected two sets of 100 images at random per class from the ImageNet dataset; one was used as the training set and the other as a tuning set during development.

We used data augmentation for the training set following Szegedy et al.\cite{Szegedy2015}. Patches of varying sizes were sampled from the image, distributed evenly between 8\% and 100\% of the image area with aspect ratio constrained to the interval $[\frac{3}{4}, \frac{4}{3}]$. For the tuning set, we sampled 320$\times$320 patches from the center of the image.

\subsection{Network architecture and training scheme}
The CNN consisted of five blocks of a convolutional layer followed by a max\=/pool layer (see Table \ref{tab:imagenetnet}). The network was optimized for 300 epochs with stochastic gradient descent, using a learning rate of $1 \times 10^{-3}$ and a mini\=/batch size of 4 images. For the streaming method, the first two blocks were streamed with tiles of 160$\times$160 pixels.

\begin{figure} 
    \centering
    {\includegraphics[width=\columnwidth]{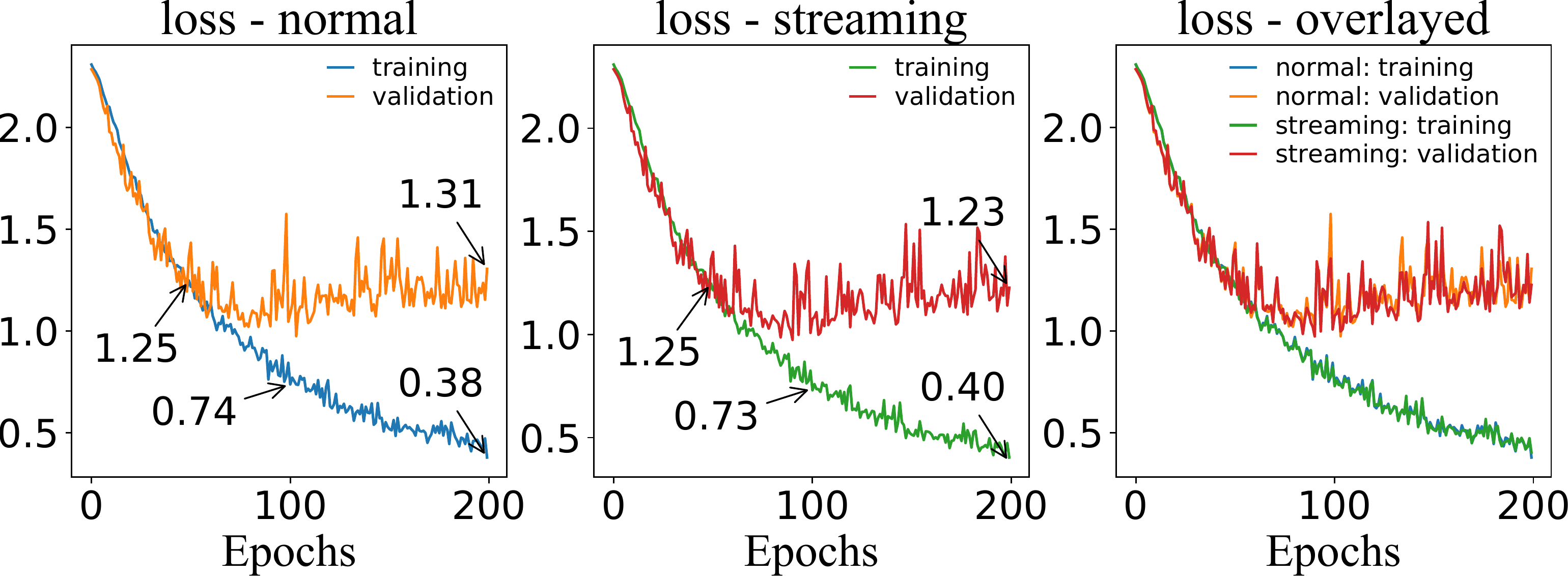}} \\
    \vspace{0.3cm}
    {\includegraphics[width=\columnwidth]{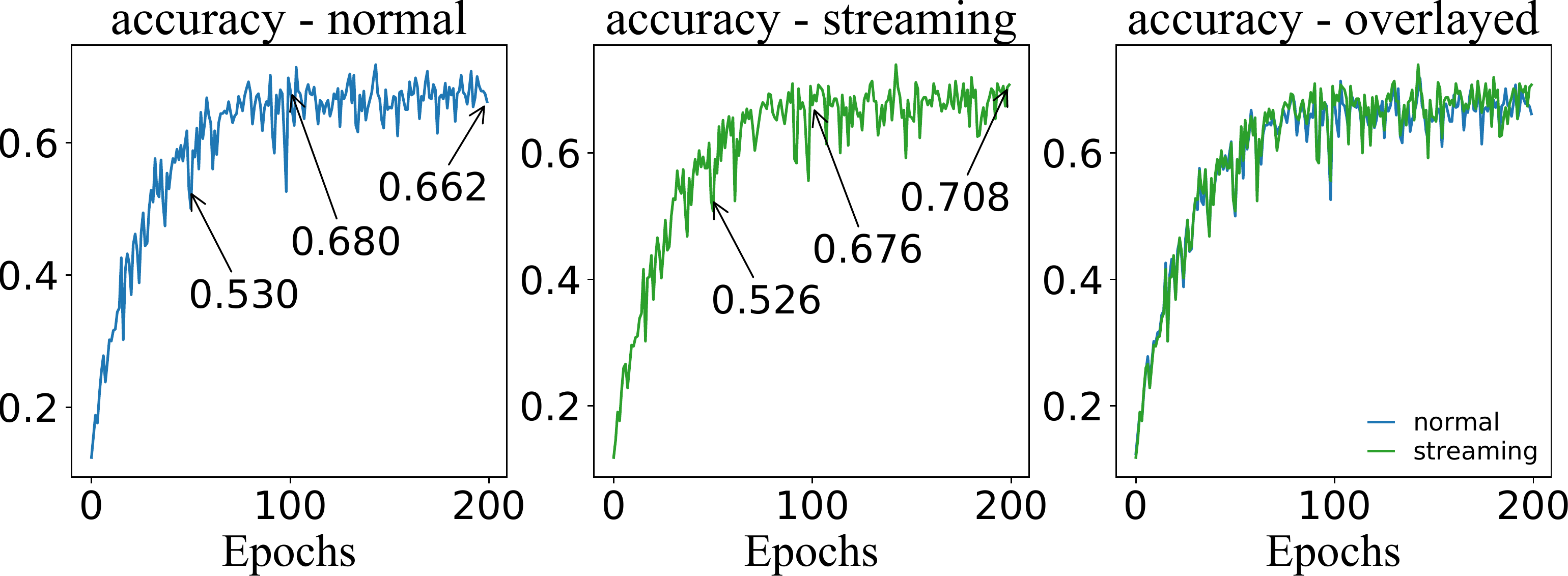}}%
    \caption{Network trained from the same initialization using conventional training and streaming (dividing the input in four tiles).}
    \label{figure:imagenette_exp}
\end{figure}

\subsection{Results on Imagenette}
The loss curves of both methods (Figure \ref{figure:imagenette_exp}) were nearly identical, which empirically shows that training with streaming performed equivalently to conventional training. Small differences are likely due to losses of significance in floating point arithmetic; these differences accumulate during training and lead to small differences in loss values in later epochs.

\section{Experiments on TUPAC16 dataset}
\label{section:tupac}
To evaluate our method on a real\=/world task, we used the publicly available dataset of the TUPAC16 challenge\cite{Veta2019}. This dataset consists of 500 hematoxylin and eosin (H\&E) stained whole\=/slide images (WSI) from breast adenocarcinoma patients. The WSIs of these patients are available from The Cancer Genome Atlas \cite{Weinstein2013} together with RNA expression profiles. The expression of 11 proliferation\=/associated genes was combined to create one objective measure for tumor growth, termed the PAM50 score\cite{Nielsen2010}. This score has no known visual substrate in the images. Thus, manual labeling is considered impossible. We set aside 98 WSIs at random for tuning the algorithm and used the remaining slides for training. Additionally, an independent evaluation was performed by the challenge organizers on the test set of 321 WSIs, of which the public ground truth is not available. The submitted predictions were evaluated using Spearman's rank\=/order correlation between the prediction and the ground truth.

\begin{table}[h]
\footnotesize
\centering
\caption{Network architecture for TUPAC16 experiments.}
\label{tab:extset}
\begin{tabular}{llll}
\toprule
\textbf{Layers} & \textbf{Kernel} & \textbf{Channels} & \textbf{Details} \\
2x 2D convolution & 3x3 & 32 & \\
2D max-pool & 2x2 & 32 \\
2x 2D convolution & 3x3 & 64 & \\
2D max-pool & 2x2 & 64 \\
2x 2D convolution & 3x3 & 128 & \\
2D max-pool & 2x2 & 128 \\
2x 2D convolution & 3x3 & 256 & \\
2D max-pool & 2x2 & 256 \\
\midrule
2x 2D convolution & 3x3 & 512 & repeated for \\
2D max-pool & 2x2 & 512 & field of view experiment \\
\midrule
2x 2D convolution & 3x3 & 512 & with BatchNormalization \\
2D max-pool & 2x2 & 512 \\
2x 2D convolution & 3x3 & 512 & with BatchNormalization \\
2D max-pool & input size & 512  & \\
Dropout (p=0.5) & & 512 &\\
Fully connected & classes & \\
\bottomrule
\end{tabular}
\end{table}

To evaluate whether CNN models can leverage and use the higher resolution information that streaming makes possible, we performed two sets of experiments. For one, we trained the same model with various image sizes (1024$\times$1024, 2048$\times$2048, and 4096$\times$4096 pixels), thus increasing input image resolution. Different networks were trained in the second set, where the depth was increased with image size (22, 25, and 28 layers for respectively 2048$\times$2048, 4096$\times$2096, and 8192$\times$8192). By also increasing the depth, the physical receptive field size before the last max\=/pool layer is kept constant (see Table \ref{tab:extset}). All networks were trained until convergence; the checkpoint with the highest Spearman's correlation coefficient on the tuning set was submitted for independent evaluation on the test set.

\subsection{Data preparation}
\begin{figure*}
\centering
{\includegraphics[width=\textwidth]{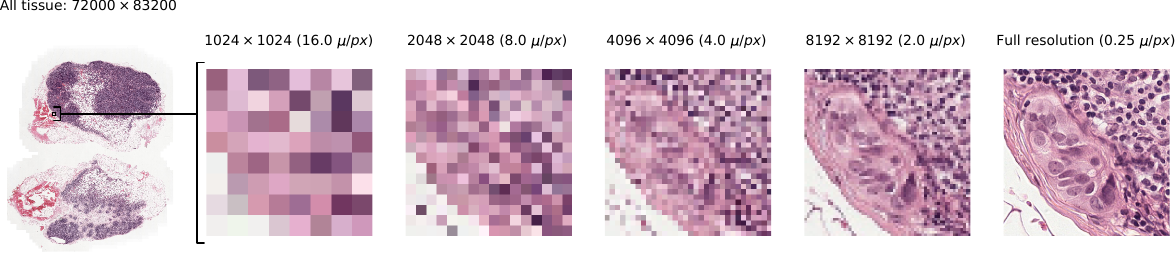}} \\
{\includegraphics[width=\textwidth]{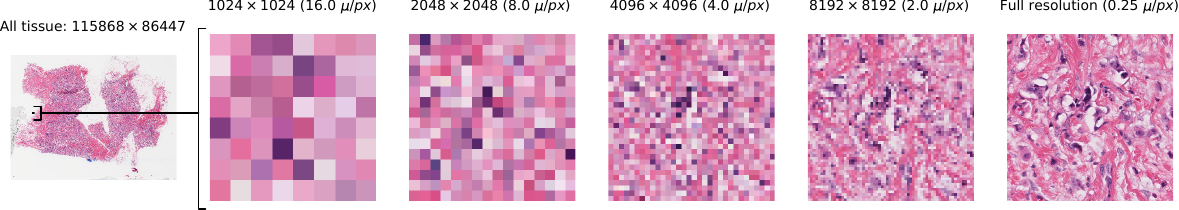}}
\caption{Resolution examples of resized whole slide images. Example slide of CAMELYON17 (top row), showing a micro\=/metastasis, and TUPAC16 (bottom row) showing stroma, illustrating the increasing detail with increasing input size.}
\label{fig:resolutions}
\end{figure*}

The images were extracted from the WSIs at image spacing $16.0\mu m$ for the 1024$\times$1024 experiments, $8.0\mu m$ for 2048$\times$2048, etc. (see Figure \ref{fig:resolutions}).  Background regions were cropped, and the resulting image was either randomly cropped or zero\=/padded to the predefined input size.

Since the challenge consists of a limited number of slides, we applied extensive data augmentations to increase the sample size (random rotations; random horizontal or vertical flipping; random brightness, contrast, saturation, and hue shifts; elastic transformations; and cutout\cite{DeVries2017}). For all experiments, the same hyperparameters and data preprocessing were used. 

\subsection{Network architecture and training scheme}
The networks (see Table \ref{tab:extset}) were trained using the Adam optimizer\cite{Kingma2015} with a learning rate of $1 \times 10^{-4}$, with the default $\beta$ parameters of $\beta_1=0.9$, $\beta_2=0.999$. We applied exponential decay to the learning rate of 0.99 per epoch. As an objective, we used the Huber loss with $\Delta=1$, also called the smooth L1 loss\cite{Girshick:2015:FR:2919332.2920125}. The mini\=/batch size was 16 images. A dropout layer with $p=0.5$ was inserted before the final classification layer. The networks were initialized following He et al.\cite{He:2015:DDR:2919332.2919814}. The images were normalized using the mean and standard deviation values of the whole training set.

Streaming was applied until the final seven layers. Since BatchNormalization breaks the local properties of chained convolutional and pooling layers, it was only used in the last part of the network. Analysis of Santurkar et al.\cite{Santurkar2018} suggests that adding only a few BatchNormalization layers towards the end of the network smooths the loss function significantly and helps optimization.

\subsection{Results on TUPAC16}
The task was evaluated using Spearman’s correlation coefficient between the prediction and the ground truth PAM50 proliferation scores. In both experiments, an improvement of the metric was seen with increasing input sizes.

\begin{table}[h]
\footnotesize
\captionsetup{justification=centering,margin=1cm}
\centering
\caption{TUPAC16: Spearman's rho on our tuning set and the independent test set}
\label{tab:tupacresults}
\begin{tabular}{llll}
\toprule
\textbf{Experiment} & \textbf{Input size} &  \textbf{Tuning set} $\bm{\rho}$ & \textbf{Test set} $\bm{\rho}$\\
\midrule
\multirow{3}{*}{Equal number of parameters} & 1024x1024 & 0.484 & 0.485 \\
& 2048x2048 & 0.624 & 0.491 \\
& 4096x4096 & 0.648 & 0.536 \\
\midrule
\multirow{3}{3cm}{Equal field of view before global max-pool (increasing depth)} & 2048x2048 & 0.624 & 0.491 \\
& 4096x4096 & 0.644 & 0.570 \\
& 8192x8192 & 0.692 & 0.560 \\
\bottomrule
\end{tabular}
\end{table}

The result of the network with the input image resolution of 4096$\times$4096 approached state\=/of\=/the\=/art for image\=/level regression with a score of 0.570. Note that the first entry of the leaderboard used an additional set of manual annotations of mitotic figures and is therefore not directly comparable to our experiments.

\begin{table}[h]
\footnotesize
\centering
\caption{TUPAC16: leaderboard}
\label{tab:tupacleaderboard}
\begin{tabular}{ll}
\toprule
\textbf{Experiment} & \textbf{Corr. coefficient} \\
\midrule
Lunit Inc., South Korea \cite{Paeng2016,Veta2019} &	0.617* \\
\textbf{Ours (4096x4096)} & \textbf{0.570} \\
\textbf{Ours (8192x8192)} & \textbf{0.560} \\
Tellez et al., 2019 \cite{Tellez2019} & 0.557 \\
Radboud UMC Nijmegen, The Netherlands \cite{Veta2019} & 0.516 \\
Contextvision, Sweden \cite{Veta2019} & 0.503 \\
Belarus National Academy of Sciences \cite{Veta2019} & 0.494 \\
The Harker School, United States \cite{Veta2019} & 0.474 \\
\bottomrule
\multicolumn{2}{c}{\footnotesize *network trained on other challenge task containing detailed annotations.}
\end{tabular}
\end{table}

\section{Experiments on CAMELYON17 dataset}
\label{section:camyleon}
CAMELYON17 was used to evaluate the streaming method on a classification task\cite{Bandi2019}. CAMELYON17 is a large public dataset and challenge to detect metastases of adenocarcinoma in breast tissue. The dataset consists of 500 labelled WSIs and 500 unlabeled WSIs, which were respectively used as the training and test sets. In the training set, for 450 slides image\=/level labels were provided, while for the remaining 50 slides dense annotations (precise delineation of the metastases) were supplied. The slides were collected from five different hospitals. The challenge differentiates three clinical relevant metastases types: macro\=/metastases ($>$ 2 mm), micro\=/metastases ($\leq$ 2.0 mm or $>$ 200 cells in a single cross\=/section), and isolated tumor cells ($\leq$ 0.2 mm or $<$ 200 cells in a single cross\=/section). We evaluate the slide level classification performance with multiple ROC analyses (one class vs. the rest).

Data preparation for this experiment was equal to the TUPAC16 challenge. We picked 90 WSIs of the challenge training set at random to be used as our tuning set.

\subsection{Network architecture and training scheme}
We used the same training schedule and underlying architecture as the TUPAC16 experiments. We altered the architecture by disabling dropout, and to reduce problems with exploding gradients in the beginning of the network, we replaced BatchNormalization with weight decay of $1 \times 10^{-6}$ and layer\=/sequential unit\=/variance (LSUV) initialization \cite{Mishkin2015}. We applied the LSUV scaling per kernel channel\cite{Krahenbuhl2015}. The mean and standard deviation per layer activation were calculated over ten mini\=/batches by keeping track of the sum and squares of the channels per tile during streaming; the reformulation of variance as $\mathbb{E}[X^2] - \mu^2$ was used to calculate the full standard deviation of ten mini\=/batches before applying LSUV.

\begin{table}[h]
\footnotesize
\centering
\caption{CAMELYON17 results on the independent test set (AUC)}
\label{tab:camresults}
\begin{tabular}{lllll}
\toprule
\textbf{Input size} & \textbf{Negative}& \textbf{ITC} &\textbf{Micro}& \textbf{Macro} \\
 & n=260 & n=35 & n=83 & n=122 \\
\midrule
2048$\times$2048 & 0.580 & 0.450 & 0.689 & 0.515 \\
4096$\times$4096 & 0.650 & \textbf{0.548} & 0.708  & 0.629 \\
8192$\times$8192 & \textbf{0.706} & 0.463  & \textbf{0.709} & \textbf{0.827} \\
\bottomrule
\end{tabular}
\end{table}

\subsection{Results on CAMELYON17}
In all cases except isolated tumor cell detection, the AUC increased with increasing resolution (See Table \ref{tab:camresults}).

\subsection{Saliency maps}
\begin{figure}[h]
  \includegraphics[width=\columnwidth]{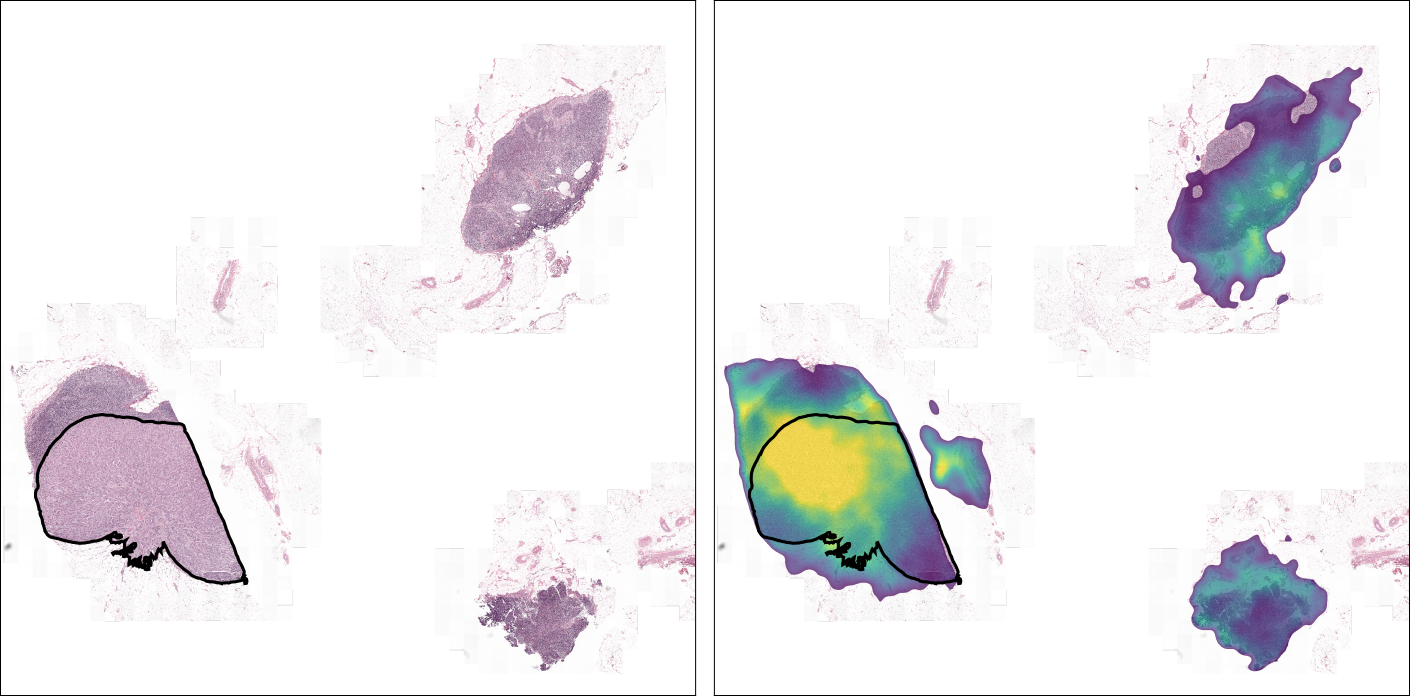}
  \includegraphics[width=\columnwidth]{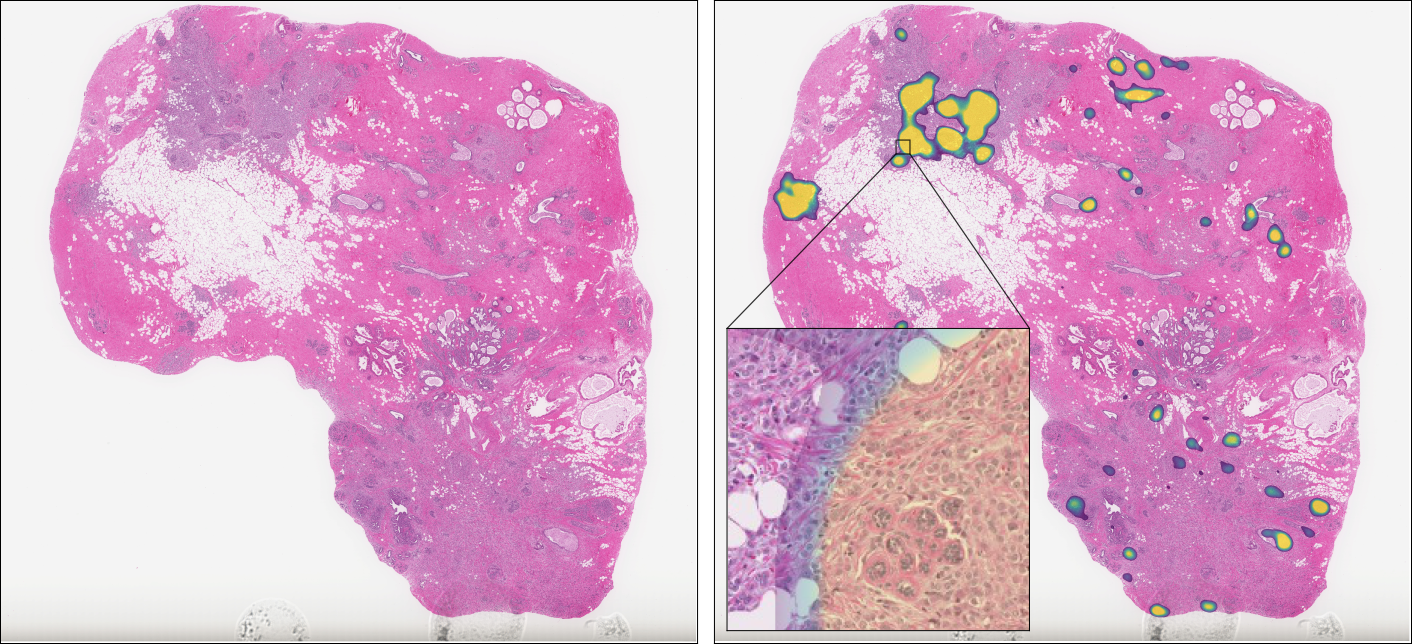}
\caption{Saliency maps for images of the tuning set of the CAMELYON17 experiment (top row) and TUPAC16 experiment (bottom row), for the networks trained on 8192$\times$8192 pixel images. The image\=/level trained CAMELYON17 network shows highlights corresponding to the ground truth pixel\=/level annotation of a breast cancer metastasis. The TUPAC16 network shows highlights in cell\=/dense and cancerous regions.}
  \label{figure:saliency}
\end{figure}

Saliency maps were created for the networks trained with the largest resolution (8192$\times$8192 pixels) according to Simonyan et al.\cite{Simonyan2013}. For better visualization on lower resolution, a Guassian blur was applied with $\sigma = 50$. Since a few gradient values can be significantly higher than others, we capped the upper gradient values at the 99\textsuperscript{th} percentile\cite{Smilkov2017}. The upper 50\textsuperscript{th} percentile was overlayed on top of the original image (See Figure \ref{figure:saliency}).

\section{Discussion and conclusion}


We presented a novel streaming method to train CNNs with tiled inputs, allowing inputs of arbitrary size. We showed that the reconstructed gradients of the neural network weights using tiles were equivalent to those obtained with non\=/tiled inputs.

In the first experiment on ImageNette, we empirically showed that the training behavior of our proposed streaming method was similar to the behavior in the non\=/streaming case. Small differences occur later in training due to loss of significance in floating\=/point arithmetic. These differences accumulated during training and lead to the small difference in loss values in later epochs. However, they do not seem to harm performance. Most modern frameworks have similar problems due to their use of non\=/deterministic operations.

The second and third experiments showed that our streaming method can train CNNs with multi\=/megapixel images that, due to memory requirements in the non\=/streaming case, would not be able to fit on current hardware. The experiment with the highest\=/resolution images ($8192 \times 8192$ pixels) would require \textasciitilde 825 gigabytes of memory per mini\=/batch if trained end\=/to\=/end in a conventional way and \textasciitilde 50 gigabytes per image.

Results on the TUPAC16 dataset (Table \ref{tab:tupacresults}) showed an increasing correlation between the prediction and the proliferation score with increasing input sizes. Our 4096$\times$4096 pixel network performed best. A jump in performance from 0.491 to 0.570 was seen from 2048$\times$2048 to 4096$\times$4096 pixels, respectively. We hypothesize that this is because tumor tissue can be discriminated from other types of tissue at these higher resolutions. However, a 8192$\times$8192 pixel input size did not further improve the performance on the test set. The nuclear details of cells at this resolution remain vague, which suggests that most of the information is still obtained from the morphology like in 4096$\times$4096 images. Higher resolutions may be necessary to further improve performance. Another explanation for the lack of improvement is the increasing difficulty for the network to find the sparse information in just 400 slides using a single label or a misrepresented tuning set due to the small provided training set. Our best result on TUPAC16 approached that of the challenge winner, who used task\=/specific information (a network trained on mitosis detection) instead of a pure regression of one label per WSI. Our method outperformed all other methods in the challenge.

Results on the CAMELYON17 dataset show improvement with increasing resolution. An exception occurs for the isolated tumor cells class; even at the highest resolution applied, the CNN was unable to differentiate isolated tumor cells. To accurately identify lesions of that size, the resolution would probably need to be increased by at least a factor of four. Furthermore, this class is also underrepresented (n=31) in the provided training set.

Using saliency maps, we visualized what the models would change on the input to make it more closely resemble the assigned class. These maps show us which parts of the image the model takes into consideration\cite{Simonyan2013}. Saliency maps of our CNNs trained on higher resolutions suggest that the networks learn the relevant features of the high\=/resolution images (see Figure \ref{figure:saliency}). The image\=/level trained CAMELYON17 network shows highlights corresponding to the ground truth pixel\=/level annotation of a breast cancer metastasis. The TUPAC16 network shows highlights in cell\=/dense regions.

The streaming method has advantages over prior work on this topic. For streaming, we do not need to alter the dataset by resizing or creating additional pixel\=/level labels (which is sometimes not possible). Also, we do not need to change the usage of the dataset like in the MIL paradigm or use compression techniques. Finally, we are not limited to specific architectural choices for our network, such as in RevNet; streaming can be applied to any state\=/of\=/the\=/art network, such as Inception or DenseNet.

While increasing input sizes and resolutions are beneficial in various tasks, there are some drawbacks. A limitation is the increase in computation time with increasing input sizes (Table \ref{tab:times}). This can be partially counteracted by dividing the batch over multiple GPUs. Due to this limitation, we did not increase resolution further in our experiments. Future research could attempt to speed up computation on the tiles, e.g., by training with mixed precision \cite{Vanhoucke2011} or depthwise separable convolutions\cite{Chollet2017}. One could also try to start with a pre\=/trained network (e.g., on ImageNet) and fine-tune for a shorter period.

Another limitation is the inability to use feature map\=/wide operations in the streaming part of the network, e.g., BatchNormalization. Future work could focus on normalization techniques that retain the local properties of the relation between the output and input of the streaming part of the network, e.g., weight normalization\cite{Salimans2016}.

Improving the performance of the high\=/resolution\=/trained networks could be a research topic of interest. In the TUPAC16 and CAMELYON17 experiments, we increased depth as we increased the input size. However, a recent work\cite{Tan2019} -- though on a maximum 480$\times$480 image size -- suggests a ``compound'' scaling rule in which the input resolution is scaled together with depth and width of the network. 

This paper focused on streaming two\=/dimensional images, but since convolutions over higher\=/dimensional data have the same local properties, one could leverage the same technique for, for example, 3D volumetric radiological images\cite{Kaggle2017}.

\section{Acknowledgements}

The authors would like to thank Erdi Calli for his help in proofreading the equations.

\small
\bibliographystyle{ieeetr}
\bibliography{library}

\section{Authors}
\subsection{Hans Pinckaers}

\includegraphics[width=0.4\columnwidth]{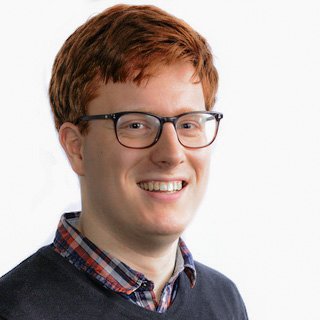}%

Hans Pinckaers is a PhD candidate in the Computational Pathology group of the Department of Pathology of the Radboud University Medical Center in Nijmegen. He studied Medicine at Leiden University Medical Center for which he obtained his MD in 2016. After his graduation he worked one year as a Pathology resident. In 2017, Hans joined the Diagnostic Image Analysis Group where he works under the supervision of Geert Litjens and Jeroen van der Laak on deep learning for improved prognostics in prostate cancer.

\subsection{Bram van Ginneken}

\includegraphics[width=0.5\columnwidth]{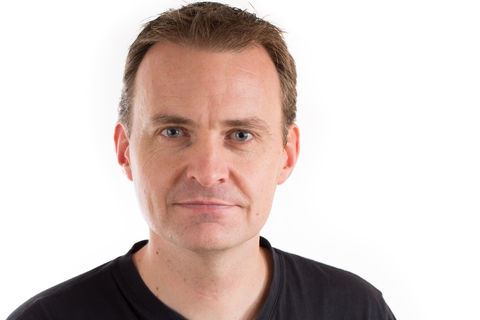}%

Bram van Ginneken is Professor of Medical Image Analysis at Radboud University Medical Center and chairs the Diagnostic Image Analysis Group. He also works for Fraunhofer MEVIS in Bremen, Germany, and is a founder of Thirona, a company that develops software and provides services for medical image analysis. He studied Physics at Eindhoven University of Technology and Utrecht University. In 2001, he obtained his Ph.D. at the Image Sciences Institute on Computer-Aided Diagnosis in Chest Radiography. He has (co\=/)authored over 200 publications in international journals. He is a member of the Fleischner Society and of the Editorial Board of Medical Image Analysis. He pioneered the concept of challenges in medical image analysis.

\small
\subsection{Geert Litjens}

\includegraphics[width=0.5\columnwidth]{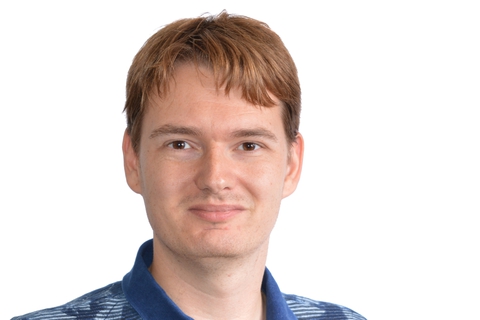}%

Geert Litjens completed his PhD thesis on ``Computerized detection of prostate cancer in multi-parametric MRI'' at the Radboud University Medical Center. He is currently assistant professor in Computation Pathology at the same institution. His research focuses on applications of machine learning to improve oncology diagnostics

\end{document}